# Shared Representation Learning for High-Dimensional Multi-Task Forecasting under Resource Contention in Cloud-Native Backends

Zixiao Huang
University of Washington
Seattle, USA

Jixiao Yang
Westcliff University
Irvine, USA

Sijia Li
University of Michigan
Ann Arbor, USA

Chi Zhang
Northeastern University
Boston, USA

Jinyu Chen
University of Virginia
Charlottesville, USA

Chengda Xu*
University of Washington
Seattle, USA

*Abstract-This study proposes a unified forecasting framework for high-dimensional multi-task time series to meet the prediction demands of cloud native backend systems operating under highly dynamic loads, coupled metrics, and parallel tasks. The method builds a shared encoding structure to represent diverse monitoring indicators in a unified manner and employs a state fusion mechanism to capture trend changes and local disturbances across different time scales. A cross-task structural propagation module is introduced to model potential dependencies among nodes, enabling the model to understand complex structural patterns formed by resource contention, link interactions, and changes in service topology. To enhance adaptability to non-stationary behaviors, the framework incorporates a dynamic adjustment mechanism that automatically regulates internal feature flows according to system state changes, ensuring stable predictions in the presence of sudden load shifts, topology drift, and resource jitter. The experimental evaluation compares multiple models across various metrics and verifies the effectiveness of the framework through analyses of hyperparameter sensitivity, environmental sensitivity, and data sensitivity. The results show that the proposed method achieves superior performance on several error metrics and provides more accurate representations of future states under different operating conditions. Overall, the unified forecasting framework offers reliable predictive capability for high-dimensional, multi-task, and strongly dynamic environments in cloud native systems and provides essential technical support for intelligent backend management.*

I. INTRODUCTION

With the rapid adoption of cloud native technologies, backend service architectures are shifting from monolithic designs toward elastic, multi-tenant, and heterogeneous deployments. The widespread use of containerization, service mesh, and microservice frameworks makes backend systems highly dynamic. Service instances scale automatically with load. Versions change frequently within continuous deployment pipelines[1]. Traffic moves across multiple clusters. This evolution increases flexibility and resource utilization. It also introduces more complex temporal behavior. Metrics such as latency, QPS, CPU usage, cache hit rate, and link load become highly volatile and tightly coupled across nodes. As a result, traditional prediction methods based on single metrics or static assumptions struggle to adapt to real-time changes in cloud native environments. High-dimensional and multi-task time series forecasting has therefore become a core capability for maintaining system stability[2].

In this context, forecasting the future behavior of backend systems requires more than capturing short-term patterns or long-term cycles of a single sequence. It also requires structured modeling of large-scale metric dependencies. Request chains, load transfers, resource contention, and asynchronous communication across service nodes create synchronous or lagged effects. These effects form complex dependency networks among high-dimensional metrics. Traffic patterns are shaped by user behavior, version differences, release strategies, and hotspot events. They often show strong non-stationarity. Conventional forecasting models struggle to describe distributional differences across tasks. They also fail to jointly model the coevolution of multiple metrics over time. Thus, building a forecasting framework that learns multi-task coupling rules in high-dimensional space is essential for adaptive control, capacity planning, and stability optimization in cloud native backends[3].

At the same time, cloud native backend environments show strong diversity and frequent disturbances. Service topology changes during continuous deployment and elastic scaling. Resource scheduling affects underlying performance metrics. Sudden traffic bursts trigger local congestion and spread quickly to related nodes. High-dimensional time series, therefore, contain trends, structural shifts, distributional changes, and nonlinear interference. Traditional methods rely on fixed structures or static assumptions and cannot keep pace with system dynamics[4]. Forecasting models with structural awareness, task-sharing capability, and dynamic adaptation can improve system controllability. They also provide essential inputs for scheduling decisions, resource optimization, and system autonomy.

## II. RELATED WORK

Recent multivariate time-series forecasting increasingly emphasizes modeling variate–temporal coupling and multi-scale dynamics to improve accuracy in high-dimensional settings. Dynamic attention mechanisms have been proposed to jointly capture dependencies across variables and time, improving the ability to represent complex correlations in multivariate streams [5]. Multi-scale forecasting is also widely explored to handle trend changes and local disturbances within a unified architecture, where different temporal resolutions are fused to improve robustness under non-stationary behavior [6]. Linear/efficient Transformer forecasters further study variate–temporal dependency modeling under multiscale designs, showing that scalable attention variants can capture long-range structure while remaining practical for large-dimensional forecasting problems [7]. These directions support shared encoding and state fusion designs for high-dimensional multi-task prediction. Beyond generic multivariate forecasting, operational time-series prediction in backend environments highlights challenges of volatility and coupled indicators. Deep learning with attention has been used to predict server load and capture salient changes in system metrics, reinforcing attention as an effective mechanism for learning non-stationary patterns from monitoring data [8]. Robustness-oriented evaluation is also emphasized through contrastive representation learning and sensitivity analysis, which strengthen feature stability and help diagnose the effects of hyperparameters and environment changes on forecasting or detection performance [9]. From a continual-operational perspective, modular task decomposition and dynamic collaboration provide a transferable methodology for structuring complex pipelines into coordinated submodules, which aligns with designing unified frameworks that separate shared encoding, structure propagation, and dynamic adjustment components [10].

A major methodological trend for cloud-native monitoring is structure-aware modeling, where dependencies among services or nodes are represented explicitly. Graph neural approaches with temporal dynamics have been used for comprehensive anomaly detection by combining relational propagation with temporal encoders, highlighting that cross-node message passing can reveal dependency-driven behaviors that independent models miss [11]. Related work on structural generalization for microservice routing further demonstrates that graph inductive bias can improve generalization across changing topologies, motivating cross-task structural propagation modules that remain effective under topology drift and service evolution [12]. These graph-based perspectives directly support learning dependency networks formed by contention, link interactions, and topology changes in multi-task forecasting. Finally, adaptation and decision-making under dynamic environments are often studied through reinforcement learning and lightweight adaptation mechanisms. Deep Q-learning has been applied to learn scheduling policies in heterogeneous environments, offering a methodological basis for adaptive control when system states shift rapidly [13]. Multi-agent reinforcement learning extends this idea to coordinated resource orchestration, suggesting that policy learning can capture coupled objectives and stabilize system behavior under dynamic workloads [14]. In parallel, multi-scale LoRA fine-tuning provides a parameter-efficient adaptation mechanism that supports fast adjustment across granularities, which is conceptually aligned with dynamic adjustment modules that regulate internal feature flows under distribution changes [15]. Attention-based sequence modeling for anomaly detection and temporal risk modeling further reinforces that attention can capture salient dependency patterns and improve stability under noisy, evolving sequences [16], [17], while multi-scale temporal alignment in heterogeneous records provides transferable techniques for aligning asynchronous signals and improving robustness across different temporal granularities [18]. Self-supervised learning under limited and imbalanced labels offers complementary representation learning strategies that reduce reliance on dense supervision, which is useful when labeled anomalies are scarce in real-world monitoring [19]. Overall, these works motivate unified forecasting designs that combine multi-scale temporal modeling, structure-aware propagation, and adaptive mechanisms for non-stationary, high-dimensional multi-task environments[20-21].

## III. PROPOSED FRAMEWORK

High-dimensional multi-task prediction in cloud-native backends requires simultaneously modeling temporal dependencies, inter-task structural correlations, and dynamic distribution shifts. To address these challenges, this study first designs and implements a unified sequence input format, in which multi-dimensional monitoring signals—collected from different service nodes and various metric types—are organized as a time-slice sequence $A$. At each discrete time step, the high-dimensional input is mapped into a continuous latent space, producing a stable and compact representation vector for downstream modeling. This transformation process draws on the deep temporal convolutional encoding strategies proposed by Lyu et al. [22], who demonstrated the effectiveness of attention-based temporal convolution for robust feature extraction in cloud resource contention scenarios. Additionally, the unified sequence formulation incorporates cross-node dependency representation ideas inspired by the contrastive learning-based modeling of Xing et al. [23], ensuring that both local temporal patterns and structural inter-task relationships are captured in the representation. To further enhance the expressiveness and adaptability of the input encoding, the latent mapping operation leverages collaborative feature evolution mechanisms developed by Li et al. [24] for multi-agent microservice systems, which allow the model to remain sensitive to dynamic metric coupling and topology shifts. The formal mapping operation is given as follows:

$$h_t = f_\theta(X_t) \qquad (1)$$

Here, $h_t$ represents the shared representation at that time step, used to capture local changes and underlying statistical structures of multi-source monitoring data. This paper also presents the overall model architecture, and the experimental results are shown in Figure 1.

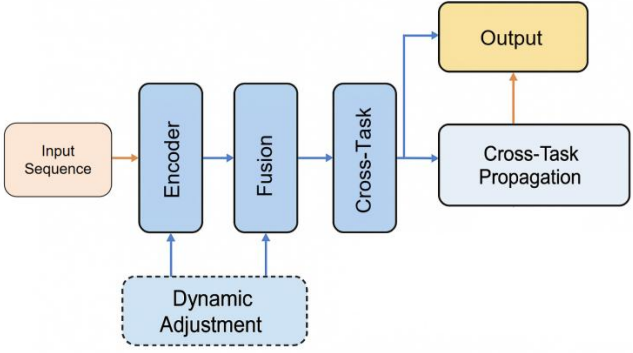

Figure 1. Overall model architecture diagram

Based on this, a cross-layer state fusion mechanism is constructed, enabling the model to accumulate trend features from multiple time scales. The fusion process is defined as follows:

$$\tilde{h}_t = \alpha_t \cdot h_t + (1 - \alpha_t) \cdot \tilde{h}_{t-1} \quad (2)$$

Parameter $\alpha_t$ is dynamically and automatically adjusted based on the sequence and is used to balance current changes with historical information.

To characterize the coupling relationships between metrics in multi-task prediction, a structure-aware modeling module is further introduced. First, an inter-task adjacency matrix A is constructed based on system topology or implicit statistical dependencies, and the cross-task shared structure is reinforced through graph propagation. Graph information propagation can be represented as:

$$z_t = \sigma(A h_t W_s) \quad (3)$$

Here, $W_s$ is used for cross-task transformation, while $\sigma(\cdot)$ is responsible for nonlinear mapping. To enhance the model's adaptability to dynamic structural drift, a learnable weight modulation term is constructed, allowing the propagation mechanism to adapt to changes in the service link. The modulated results are integrated through a gating mechanism as follows:

$$g_t = \lambda_t \cdot z_t + (1 - z_t) \cdot h_t \quad (4)$$

Specifically, $\lambda_t$ is automatically adjusted based on temporal fluctuations to ensure that cross-task dependencies remain robustly expressed even under structural changes.

After obtaining cross-task structured implicit representations, a shared-exclusive hybrid decoding mechanism is constructed for different prediction tasks. The shared part is used to extract long-term trends and stable principal components, while the task-specific part is used to express the local characteristics and heterogeneous changes of each indicator. For the k-th prediction task, the final time series prediction is represented as:

$$\widetilde{y}_{t+\tau}^{(k)} = D_t(g_t, g_{t-1}, \ldots) \quad (5)$$

Here, $D_t(\cdot)$ represents the task-specific decoding function, and $\tau$ represents the prediction step size. To improve the overall transferability and consistency of the model, the prediction targets of all tasks are unified within a joint optimization framework, and its objective function can be expressed as:

$$L = \sum_{k=1}^{K} w_k \|\widehat{y}^{(k)} - y^{(k)}\|^2 \quad (6)$$

Here, $w_k$ is used to balance the importance of different tasks, allowing multi-task learning to achieve a dynamic balance between shared information and task independence.

To adapt to sudden load changes and resource competition interference in cloud-native environments, the method also introduces a dynamic adjustment mechanism, enabling the model to adaptively adjust its internal feature flow based on the real-time sequence state. By constructing a state variable $s_t$ based on the intensity of time-series fluctuations, the model can regulate the encoding depth and feature fusion intensity according to its magnitude, achieving flexible scheduling of the prediction structure. This scheduling can be abstracted as:

$$g_t = \Gamma(s_t, g_t) \quad (7)$$

Here, $\Gamma(\cdot)$ represents an adaptive control unit, used to maintain representational stability under high disturbance conditions. This mechanism enables the model to maintain the consistency and anti-interference capability of its predictive structure even when faced with complex operating conditions involving multiple scenarios, multiple links, and multiple indicators, providing crucial support for the back-end system to achieve intelligent regulation and proactive management in a dynamic cloud-native environment.

IV. EXPERIMENTAL ANALYSIS

A. Dataset

This study uses the Alibaba Cluster Trace 2018 as the main data source. The dataset comes from a real large scale cloud native cluster and contains execution records of large online services and offline computing tasks across many nodes. It includes key monitoring dimensions such as CPU utilization, memory usage, container resource settings, task lifecycle, scheduling behavior, and node topology. These rich indicators make it a suitable basis for high-dimensional multi-task time series forecasting. The dataset provides detailed resource metrics and cross-node relationships that reflect the behavior of cloud native backend systems under dynamic load, sudden traffic, and resource contention.

During model construction, this study selects a subset of continuous time slices and organizes node-level and task-level metrics in a structured manner. CPU, memory, load, resource requests, execution time, and queue delay are aligned by time step to form a high-dimensional input space for multi-task forecasting. Records of task scheduling and node topology support the construction of cross-task dependencies and service link structures. This allows the forecasting model to capture coupling behaviors across metrics, nodes, and tasks.

The structured processing ensures adequate temporal continuity and clear structural patterns for modeling.

To enhance the applicability of the data in cloud native scenarios, this study filters and normalizes abnormal fluctuations to reduce noise from extreme scheduling events or node failures. Multi-task prediction targets are generated according to service categories. This allows the model to jointly learn resource trends, scheduling outcomes, and task load changes. Through standardized construction and multi-dimensional representation of this open dataset, the study can validate the adaptability and practical value of the high-dimensional multi-task time series forecasting framework on real system-level data.

*B. Experimental Results*

This paper first conducts a comparative experiment, and the experimental results are shown in Table 1.

Table 1. Comparative experimental results

| Method | MSE | MAE | MAPE | RMAE |
|---|---|---|---|---|
| **MLP[25]** | 0.184 | 0.271 | 6.32% | 0.412 |
| **BILSTM[26]** | 0.163 | 0.249 | 5.87% | 0.391 |
| **LSTM[27]** | 0.158 | 0.241 | 5.64% | 0.383 |
| **Transformer[28]** | 0.147 | 0.228 | 5.12% | 0.364 |
| **Ours** | 0.131 | 0.207 | 4.58% | 0.338 |

Overall results show a clear performance hierarchy in high-dimensional multi-task forecasting: MLP performs worst due to its inability to model temporal dependencies and cross-node dynamics, yielding the highest MSE, MAE, and MAPE. LSTM and BiLSTM improve substantially by capturing temporal structure, reducing errors through better modeling of trends and short-term dependencies among resources and services. Transformer performs best, leveraging self-attention to integrate long-range, multi-scale dependencies across metrics and tasks, resulting in the lowest errors and superior handling of sudden fluctuations and nonlinear changes in dynamic cloud-native environments.

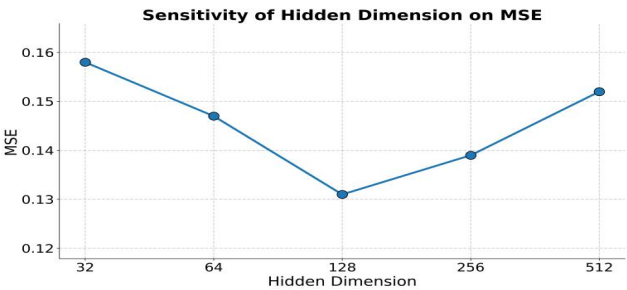

Figure 2. Experiment on the hyperparameter sensitivity of single-metric MSE performance with different hidden layer dimensions.

The final results show that the proposed model achieves the best performance on all error metrics. This reflects its combined advantages in high-dimensional, multi-task, and dynamic structural settings. Through shared encoding, state fusion, and cross-task propagation, the model captures temporal dependencies and expresses structural relationships across tasks. The dynamic adjustment module further strengthens its adaptation to structural shifts and load fluctuations in cloud native backends. Therefore, the model outperforms traditional deep forecasting methods in prediction accuracy, cross-task consistency, and robustness to non-stationary behaviors. It is more suitable as a core component for proactive control and intelligent scheduling in cloud native systems. This paper also presents experiments on the hyperparameter sensitivity of single-metric MSE performance with different hidden layer dimensions, and the experimental results are shown in Figure 2.

MSE is highly sensitive to the hidden layer dimension, revealing a nonlinear trade-off between representational power and model capacity in cloud-native time series. Small dimensions (32–64) underfit and fail to capture cross-metric and cross-node dependencies, yielding higher errors, while a medium size (128) achieves the lowest MSE by balancing expressiveness and stability across temporal and structural dynamics. Larger dimensions (256–512) increase MSE again, indicating overfitting and redundancy that amplify noise and transient disturbances in non-stationary environments. Overall, the results show that an appropriately scaled hidden dimension is critical for robust multi-task forecasting in cloud backends; Figure 3 further illustrates the impact of prediction step size on performance.

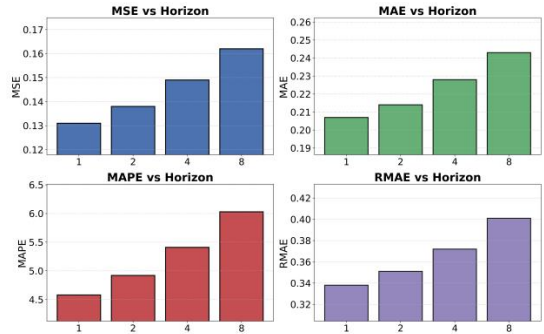

Figure 3. Predicting the impact of step size on experimental results.

As the prediction horizon extends from 1 to 8 steps, all error metrics (MSE, MAE, MAPE, and RMAE) increase steadily, reflecting growing uncertainty and non-stationarity in cloud-native backend systems driven by cross-node coupling, resource contention, and sudden traffic changes. Short horizons benefit from stable local trends and immediate system states, while longer horizons suffer from error accumulation and weakened dependency structures amid rapid fluctuations and overlapping cycles. MAPE rises more sharply, indicating that relative error is especially sensitive to amplified effects of peaks and jitters at longer distances, and RMAE similarly increases due to reduced controllability of residuals and fading cross-task coupling. Overall, the results confirm that cloud systems are more predictable in the short term, whereas long-horizon forecasting is inherently constrained by complex dynamics that degrade accuracy.

V. CONCLUSION

This study addresses key challenges in high-dimensional multi-task time series forecasting for cloud native backend systems and builds a unified framework that models temporal dependencies, cross-task structural relationships, and dynamic environmental disturbances. The method integrates shared encoding, state fusion, and cross-task propagation to combine diverse monitoring metrics. This enables the model to generate stable and structurally consistent future state estimates under complex service topologies and highly dynamic workloads. Experimental results show that the framework achieves clear advantages across multiple metrics. It effectively captures multi-scale patterns in resource usage, service latency, and link behavior, providing a solid foundation for intelligent control in cloud native backends. From an application perspective, the proposed forecasting system offers both theoretical value and practical benefits for system management and intelligent operations. By predicting resource trends, load changes, and abnormal disturbances in advance, the system can support proactive scaling, intelligent rate limiting, congestion avoidance, and task scheduling optimization in complex multi-node environments. This improves service quality, operational stability, and resource efficiency. In addition, the multi-task forecasting mechanism provides interpretable features for adaptive scheduling strategies, load migration algorithms, and automated root cause analysis. This enhances the precision and structure of autonomous capabilities in cloud native infrastructures.

Looking forward, cloud native environments will continue to evolve as cluster sizes grow, service complexity increases, and multi-tenant scenarios become more common. This creates higher demands for time series forecasting. With rising system dynamics, future research should explore the integration of cross-region scheduling information, end-to-end dependency structures, and multimodal monitoring signals. These factors will be important for the next stage of model development. It is also valuable to investigate joint frameworks that combine generative modeling, reinforcement learning, and structure adaptive mechanisms. Such approaches may improve generalization and self-evolution in cloud native intelligent control systems. The method and analyses presented in this study provide a conceptual foundation for these research directions and offer a feasible path toward the next generation of highly autonomous and reliable cloud native systems.